\pdfoutput=1 
\documentclass[10pt,twocolumn,letterpaper]{article}
\usepackage{cvpr}              
\usepackage{graphicx}
\usepackage{amsmath}
\usepackage{amssymb}
\usepackage{booktabs}

\usepackage{algpseudocode}
\usepackage{float} 
\usepackage[pagebackref,breaklinks,colorlinks]{hyperref}

\usepackage[capitalize]{cleveref}
\crefname{section}{Sec.}{Secs.}
\Crefname{section}{Section}{Sections}
\Crefname{table}{Table}{Tables}
\crefname{table}{Tab.}{Tabs.}

\def\cvprPaperID{14} 
\def\confName{CVPR}
\def\confYear{2022}

\newcommand{\p}{\ensuremath{{\cal P}}}

\newcommand{\cv}{\ensuremath{{\cal V}}}

\newcommand{\R}{\ensuremath{{\mathbb R}}}

\newcommand{\bS}{\ensuremath{{\mathbb S}}}

\newcommand{\E}{\ensuremath{{\mathbb E}}}

\newcommand{\beq}{\begin{equation}}
\newcommand{\eeq}{\end{equation}}
\newcommand{\beqa}{\begin{equation}\begin{aligned}}
\newcommand{\eeqa}{\end{aligned}\end{equation}}
\newcommand{\brmk}{\begin{rmk}}
\newcommand{\ermk}{\end{rmk}}
\newcommand{\partref}[1]{\hbox{(\csname @roman\endcsname{\ref{#1}})}}

\begin{document}

\title{ImageSig: A signature transform for ultra-lightweight image recognition}

\author{Mohamed R. Ibrahim$^1$ \\
$^1$The Alan Turing Institute\\
London, UK\\
{\tt\small mibrahim@turing.ac.uk}
\and
Terry Lyons$^{1,2}$\\
{$^2$ Mathematical Institute, Oxford University}\\
{Oxford, UK}\\
{\tt\small terry.lyons@maths.ox.ac.uk}
}

\maketitle

\begin{abstract}
This paper introduces a new lightweight method for image recognition. ImageSig is based on computing signatures and does not require a convolutional structure or an attention-based encoder. It is striking to the authors that it achieves: a) an accuracy for 64 X 64 RGB images that exceeds many of the state-of-the-art methods and simultaneously b) requires orders of magnitude less FLOPS, power and memory footprint. The pretrained model can be as small as 44.2 KB in size. ImageSig shows unprecedented performance on hardware such as Raspberry Pi and Jetson-nano. ImageSig treats images as streams with multiple channels. These streams are parameterized by spatial directions. We contribute to the functionality of signature and rough path theory to stream-like data and vision tasks on static images beyond temporal streams. With very few parameters and small size models, the key advantage is that one could have many of these "detectors" assembled on the same chip; moreover, the feature acquisition can be performed once and shared between different models of different tasks - further accelerating the process. This contributes to energy efficiency and the advancements of embedded AI at the edge.
\end{abstract}

\section{Introduction}
\label{sec:intro}

Image recognition represents the backbone for all vision tasks in which its accuracy and efficiency are seminal for tasks such as classification, object detection, or semantic segmentation. Advances in image recognition have been achieved relying on very deep convolution models \cite{ResNet50, xception,inceptionv3,efficientnet}. Most recently, Attention-based models such as Vision Transformers (ViT)\cite{vit} have also shown strong progress for vision tasks at scale when trained in large datasets. While both approaches (Convolution-based and attention-based models) have yielded high accuracies in most benchmark datasets (\eg ImageNet \cite{imagenet}), this performance often comes at the expense of the model weights, the number of parameters and subsequently the time and resources needed for training and inference. Even with mobile-friendly methods (\eg MobileNets \cite{mobilenets,mobilenetv2, mobilenetv3}) and post-training quantization techniques \cite{quant_1,quant_2}, the number of parameters still exceeds millions. This makes it a challenge for embedding AI for microcontrollers or devices with minimal computational resources when implementing an entire pipeline of multiple vision models. 

In this paper, we introduce a new method, called ImageSig, that is suitable for training lightweight vision models. Based on rough path theory \cite{rough_path_book,rough_path_paper}, we treat a single image as a stream of paths over a virtual time, whereas pixels evolve instantaneously over its spatial attributes. The architecture of ImageSig relies on extracting a unique signature from these paths that can be trained directly by a fully connected layer for classification without the need for extracting features based on expensive operations such as 2D convolution layers or attention-based structures. We reveal that ImageSig can provide a faster and efficient way for training deep models with very few parameters and model footprint that is often less than 1 MB without quantization and drops as low as 44.2 KB after quantization.


\section{Related work}
There are three domains that are relevant to the stated issue. We have reviewed the relevant work, and here we highlight the key challenges where further research is needed.

\textbf{Lightweight convolution models:}  
Training fast, efficient, and lightweight convolution models is ongoing research \cite{mobilenets,mobilenetv2,mobilenetv3}. There are several approaches that are of particular interest to scholars and have been intensively addressed in the literature.  First, lightweight models have been achieved by training smaller networks of a larger one of the same family (\eg ResNet18 vs ResNet 101 \cite{ResNet50}). However, this approach often comes at the expense of accuracy when compared to the larger networks of the same family. The second approach is by training a larger network and distillates its knowledge to a smaller network with fewer parameters \cite{knowledge_dist}. Transferring knowledge from teacher to student architecture was key in many applications. However, on others, it remains an expensive solution for training a given task. Nevertheless, the weights of a student network do not often show drastic changes in the overall number of parameters. Instead, it is often reported as a slightly lighter version of a given teacher network. Third, by relying on various factorization-based methods (\eg separable convolutions \cite{xception}, dimension-wise convolutions \cite{dicenet}), various mobile-friendly methods have been achieved for lightweight vision tasks such as MobileNets \cite{mobilenets,mobilenetv2,mobilenetv3}, ShuffleNetV2 \cite{shufflenet},  MNASNet \cite{mnasnet}, and ESPNetv2 \cite{espnetv2}, which have been proven to be more efficient in comparison to deep convolution models. Yet, these models have been reported with millions of parameters for training. Nevertheless, this \emph{lightweight} comparing to deeper models, comes at the expense of accuracy. Last, post-training pruning \cite{prune} and quantization \cite{quant_1,quant_2,quant_3,quant_4} have shown success in improving model efficiency through reduced precision arithmetic without compromising its overall utility and accuracy.

\textbf{Transformers, patch and position embeddings:} Applying transformers \cite{transformer} to visual perceptions has opened the door for understanding vision tasks beyond the need for a convolutional structure.  In Vision Transformers (ViT) \cite{vit}, a given Image is subdivided into small and fixed-size patches in which their positions are spatially registered by embeddings and then feedforward to a transformer encoder.  Images are transformed into patch embeddings, in which a given 2D image $X\in\mathbb{R}^{H\ X\ W\ X\ C}$ is reshaped and flatted to 2D patches    $X_P\in\mathbb{R}^{N\ X\ (P^2 X C)}$ where $(H,W)$ is the resolution of the given image, $C$ is the number of channels, $(P,P)$ is the resolution of the given patch, and N represents the number of patches, given that $N=HW/P^2$.  In addition to the patch embeddings, a learnable 2D position embedding is added to preserve the positional information of each patch without inductive bias; the spatial relations among the patches need to be learned from scratch. While this approach has achieved 88.55\% in the ImageNet dataset, it requires a large network with millions of parameters and expensive computational resources (30 days on a TPUv3 with 8 cores) trained exclusively on large datasets. To allow ViT to be trained on small-sized datasets, a Compact Convolution Transformer (CCT) \cite{cct} is introduced. Unlike ViT, the CCT model introduced a convolution block, in which image patching and embeddings are implemented to its output to reduce the latent representation of the transformer where $X_0=MaxPool(ReLU(Conv2d(X))$. This allows the model to learn from relatively smaller datasets based on the inherited inductive bias and spatial relationships of the pixels. This also allows changing the resolution of the given image without increasing the number of parameters of the model. However, this remains at the expense of the sequence length and, subsequently, the computational resources. On the other hand, a Perceiver \cite{perceiver} is introduced to leverage an asymmetric attention mechanism to iteratively map a byte array (at a pixel level) as an input to a smaller latent array. This approach allows the transformer to handle a large input size of various modalities by uncoupling the depth of the transformer and the exponential size of the latent representation of the transformer based on the input resolution. While there are also several modifications of transformer models that leverage one or more of the aforementioned concepts \cite{chen2021outperform,tolstikhin2021mixer,steiner2021augreg, perceiver_IO, convit, mobilevit,delight}, the challenges remain on how to handle large size byte arrays to the latent representation of a given transformer encoder with respect to the number of the parameters and subsequently computational resources.

\textbf{Signature-based methods:} Based on rough path theory, a signature is a unique summary of a given path \cite{review_sig,rough_path_book,rough_path_paper,kidger2021signatory,sig_5}. What makes signature a robust method for describing a path is its invariant characteristics to the reparameterisation of the given path. This provides a natural feature of linear functionals that only capture the key elements of the given path by mapping the orders of the information of the stream instead of mapping precisely the location of the path at each instance \cite{rough_path_paper}. Transforming datasets into streams is a modelling issue that requires assumptions regarding to how the system is altered once new information arrives \cite{review_sig}. There are no signature-based methods for image recognition. However, Signature transforms with deep learning have been introduced \cite{deep_sig}, in which different scholars have utilised it for various machine learning applications. For example, a deep model for recognising online handwritten characters from temporal paths has been introduced  \cite{sig_handwriting}. Based on pose estimation, a signature transform has been applied to recognise human actions based on their landmarks \cite{sig_landmark}. This paper contributes to extending rough path theory and signature transform directly to vision tasks and stream-like data regardless of the presence of an explicit path.


\section{Methodology}
\subsection{Image signature}

In this research, we contribute to how signature methods can be applied to stream-like data. Nevertheless, we assume that signature transforms can be perceived as the equivalent of spatial embeddings that can offer orders for a byte array (at a pixel level) while reducing its dimensionality. 
We assume that $V$ is a $d$-dimensional vector (Banach) space of basis $\mathcal{B}={v_1,...v_d}$ and defined based on tensor algebra as: 

\beq
	\label{tv}
T(V){\colon=\ \bigoplus_{k=0}^\infty}\ V^{\bigotimes k}	
\eeq
given that $V^{\bigotimes0}:=\R$. We denote the projection of $T(V)\rightarrow V^{\bigotimes n}$ by $\pi_n$ where $n\in\mathbb{N}_0$ and the truncated tensor algebra is defined as:
\beq
	\label{tv2}
T^{(n)}(V){\colon= \bigoplus_{k=0}^n}\ V^{\bigotimes k}  
\eeq



An image is transformed into paths by imagining a given 2D image $X\in\mathbb{R}^{H\ X\ W\ X\ C}$ as patches of streams $\Omega_v \subset V$  of discontinuous paths $\gamma:\ [a,b] \to V$ where a given pixel $(i)$  evolves, in magnitude, from left to right over a virtual time to the adjacent pixel given that $i\in W$. In this order, changes occur instantaneously given that signature is invariant to reparameterisations. The channels of a given image $C$ represents the depth of a given path (See Fig. \ref{fig:method_fc}).  The truncated signature of $\gamma$  at a given depth $N$ is defined as:
\beq
	\label{siq1}
S_{a,b}(\gamma) = \bigoplus_{n=0}^N S_{a,b}^n(\gamma),
\eeq
given that:
$\ S_{a,b}^n(\gamma)=\ \frac{1}{n!}\left(\gamma_b-\gamma_a\right)^{\bigotimes n}$

So the signature transform given that ${Sig}^N=\ \mathbb{S}(\mathbb{R}^d)\rightarrow\prod_{n=1}^{N}{(\mathbb{R}^d)}^{\bigotimes n}$ is defined as:




\beq
       \resizebox{0.9\hsize}{!}{$
		Sig^N (X) =
		\idotsint\limits_{0 < t_1 < 
		\ldots < t_n < 1}		
		\left( \frac{df}{dt} (t_1)\right) \bigotimes \ldots \bigotimes
		\left( \frac{df}{dt} (t_n)\right) ((24dt_1)  
		\ldots (24dt_n))_{1\le n\le N}$}
\eeq

And the log signature of $\gamma$ is defined as: 
\beq
	\label{log_sig}
log S_{a,b}(\gamma){= \bigoplus_{n=0}^N}{\frac{{(-1)}^{n-1}}{n}\left(\ {\hat{S}}_{a,b}^n(\gamma)\right)}^{\bigotimes n}
\eeq
where $S_{a,b}^0(\gamma)=1$ 
and ${\hat{S}}_{a,b}(\gamma){:=\ \bigoplus_{n=1}^N}\ S_{a,b}^n(\gamma)$.

The collection of paths $\mathcal{N}\subset\mathcal{V}_1[a,b];V$, given that  $N \in H$  can be summarised as $\delta_\mathcal{N}=\frac{1}{\left|\mathcal{N}\right|}\sum_{\gamma\in\mathcal{N}\ }\delta_\gamma$. 
The \emph{expected signature} of the collective paths that represents an image is the map $\bS : \p \cv_1 ( [a,b] ; V) \to T(V)$ in a probability (Borel) space and is computed as: 
\beq
	\label{expect_sig_def}
		\bS (\mu) := \E_{\mu} \left[ S_{a,b}(X) \right]
		= \bigoplus_{n=0}^{N}   
		\E_{\mu} \left[ S^n_{a,b}(X) \right]
\eeq

We highlight that the proposed image signature summarises and reduces the dimensionality of a given image $X$ into $\hat{X}\in\mathbb{R}^{H\ X\ Q}$, where  $Q= \sum_{n=1}^{N}C^n$. It is worth mentioning that we also considered looking at a given image as two-directional streams of paths, horizontally and vertically, in which we concatenated both outcomes where  $\hat{X}\in\mathbb{R}^{2  X (H\ X\ Q)}$ and we refer to this method as \emph{two-directional streams}. Second, the aspect ratio of a given image can be taken into consideration without the need for finding an equal dimension for both $H$ and $W$. Last, while the computed signature is variant to $W$, the dimensions of $\hat{X}$ is invariant to $W$. This feature allows training a model in a dataset of images of various aspect ratios, providing that $H$ remains constant.

\subsection{Signature augmentation}
While the computed signature is invariant to reparameterisation, it remains variant to change in colour, rotation, and displacement of a given image. Nevertheless, the order in which images are transformed is a modelling choice. Accordingly, similar to image augmentation techniques \cite{data_augmentation,data_augmentation2,RandAugment}, we applied several data augmentation techniques such as adding Gaussian noise, brightness, horizontal flipping, rotation, and colour invert to images before computing their respective signature without altering a given class of an image.

\subsection{Encoder architecture}
After computing image signature, we experimented with three \emph{minimalist architectures} to show the robustness of ImageSig; one architecture is based on using only a single FC layer to classify a given image,  the second comprises a convolution encoder, and the third is based on a transformer encoder. 


\begin{figure*}
\begin{center}
\includegraphics[width=0.8\linewidth]{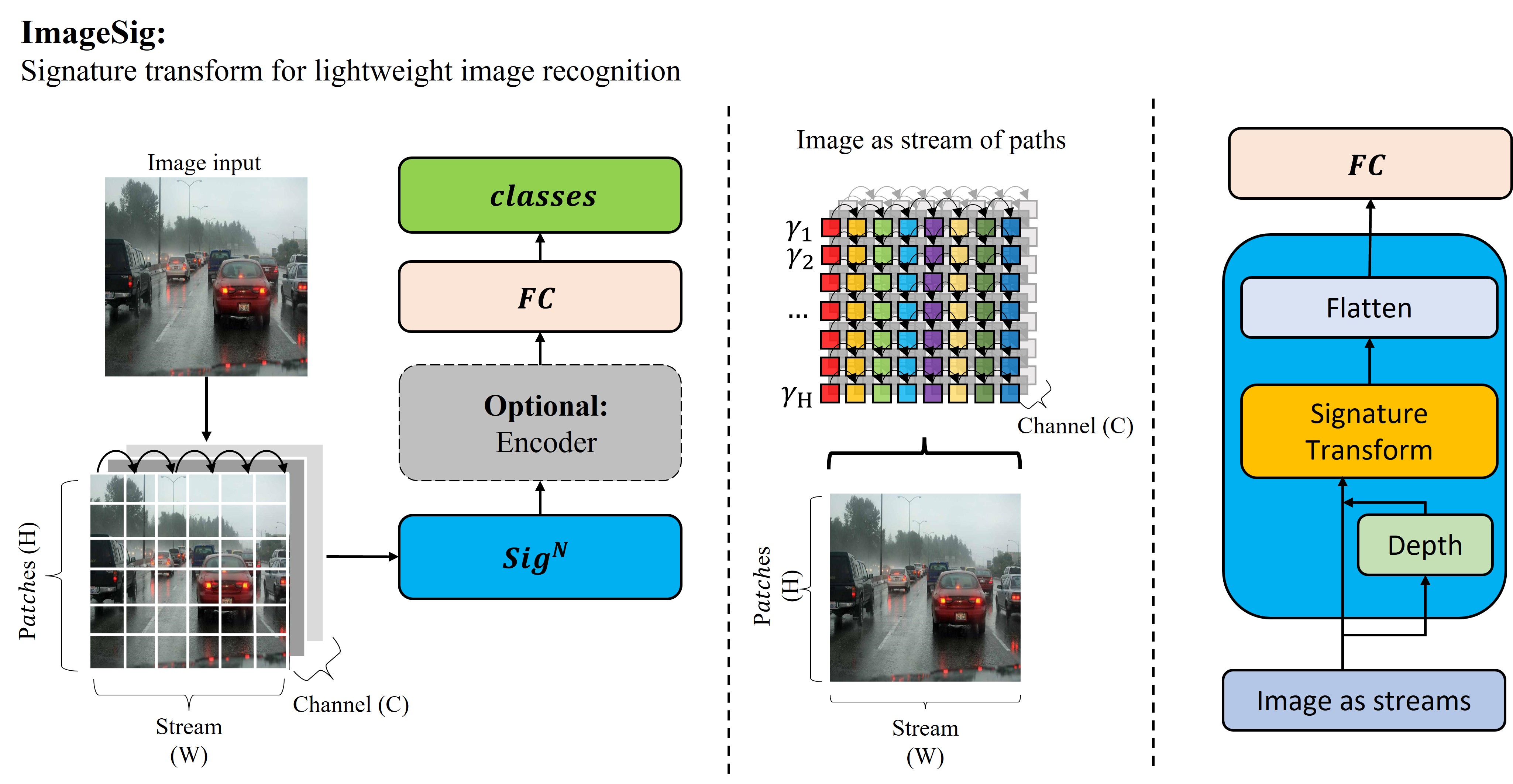}
\end{center}
   \caption{Proposed method: ImageSig.}
\label{fig:method_fc}
\end{figure*}


\textbf{A single FC-based model:} 
This model represents the simplest approach for training a vision model. After computing the image signature and flattening its output, the architecture of the model relies only on a single Fully connected layer (FC) of 50 neurons activated by a ReLU function, followed by an output softmax layer of neurons equal to the number of classes in a given dataset. Fig. \ref{fig:method_fc} shows the overall architecture of ImageSig.

\textbf{A 1D convolution-based model:}
As shown in Fig. \ref{fig:method_fc}, an encoder can be added to extract features and downsample the computed signature before the FC layer. The objective of this architecture is to evaluate the effect of adding a convolution block on the overall performance,  bearing in mind the trade-off between performance and the overall weight of the model.  Accordingly, we used a simple convolution block comprised of two CNN 1D layers, in which each layer is followed by a Max-pooling 1D layer. After the last pooling layer, the model is flattened and feedforward to a single FC layer similar to the aforementioned architecture. 

\textbf{Attention-based model:} 
We also feedforward the computed signature (without the Flatten layer) to a transformer encoder similar to the one implemented in CCT model \cite{cct}, but without any patch or position embeddings. And before the multi-head attention layers, we used CNN1D layers (similar to the aforementioned architecture) instead of the CNN2D layers.

\subsection{Objective loss}
We penalise the introduced method based on a cross-entropy coupled with a focal loss \cite{focal_loss} to account for class imbalance for data sets of skew representation for each class. For binary classification, given that
$y \in \{0, 1\}$
and $\hat{p} \in [0, 1]$, the objective loss $(L)$ is defined as: 
\beq
	\label{loss}
L(y, \hat{p}) = -\alpha y \left(1 - \hat{p}\right)^\gamma \log(\hat{p})
                - (1 - y) \hat{p}^\gamma \log(1 - \hat{p})
\eeq
where $\gamma$ represents the focusing parameter that postulates the confidence level for the contributions of correct predictions to the overall loss (the higher the $\gamma$, the higher the rate for down-weighting examples that are easy to classify) and $\alpha$ represents a hyperparameter that specifies the trade-off between recall and precision by increasing or decreasing the weight of errors for the positive class (when $\alpha=1$ is the same as no weighting).


\section{Experiments}
We report on over \emph{40 ImageSig models}, in addition to base models, trained on \textbf{four datasets} with different encoder architectures and various hyperparameters which have been deployed for different hardware (\eg Raspberry Pi).

\subsection{Datasets}
\begin{figure}
\begin{center}
\includegraphics[width=0.75\linewidth]{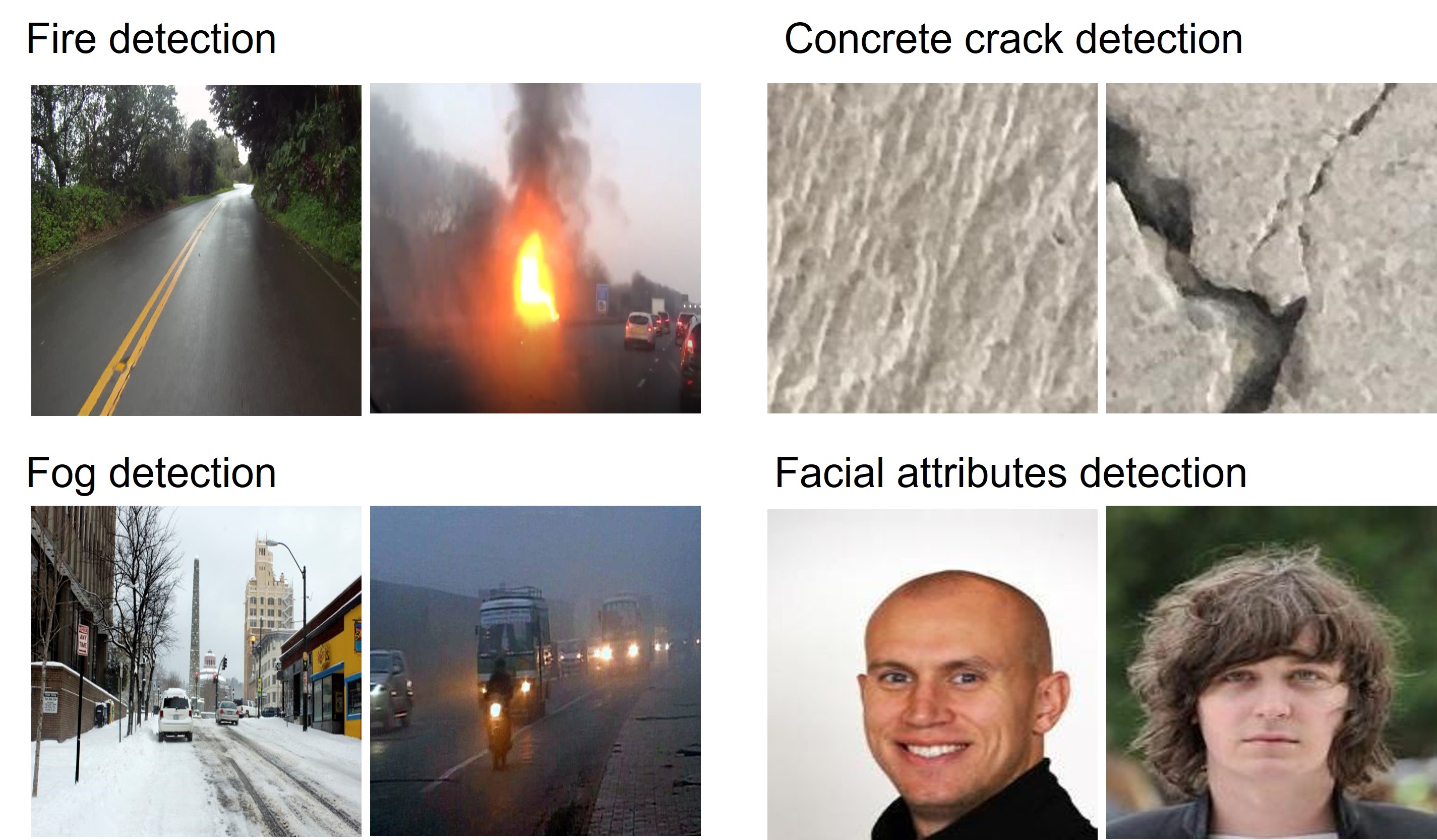}
\end{center}
   \caption{Image samples from the four datasets. }
\label{fig:dataset_sample}
\end{figure}

We aim to apply our introduced method to applications to the real-world (\eg anomaly detection from images). Accordingly, we have applied ImageSig to both benchmarked and custom datasets that are relatively small in size and imbalanced in classes which are common issues when it comes to practical applications (See Fig. \ref{fig:dataset_sample}). On the other hand, it is often under-reported how the introduced state-of-the-art methods perform in real-world applications beyond benchmarked datasets. Furthermore, we also aim to show how our method performs when it compares, for instance, to transformers that usually require large datasets for training.

\begin{figure}
\begin{center}
\includegraphics[width=0.75\linewidth]{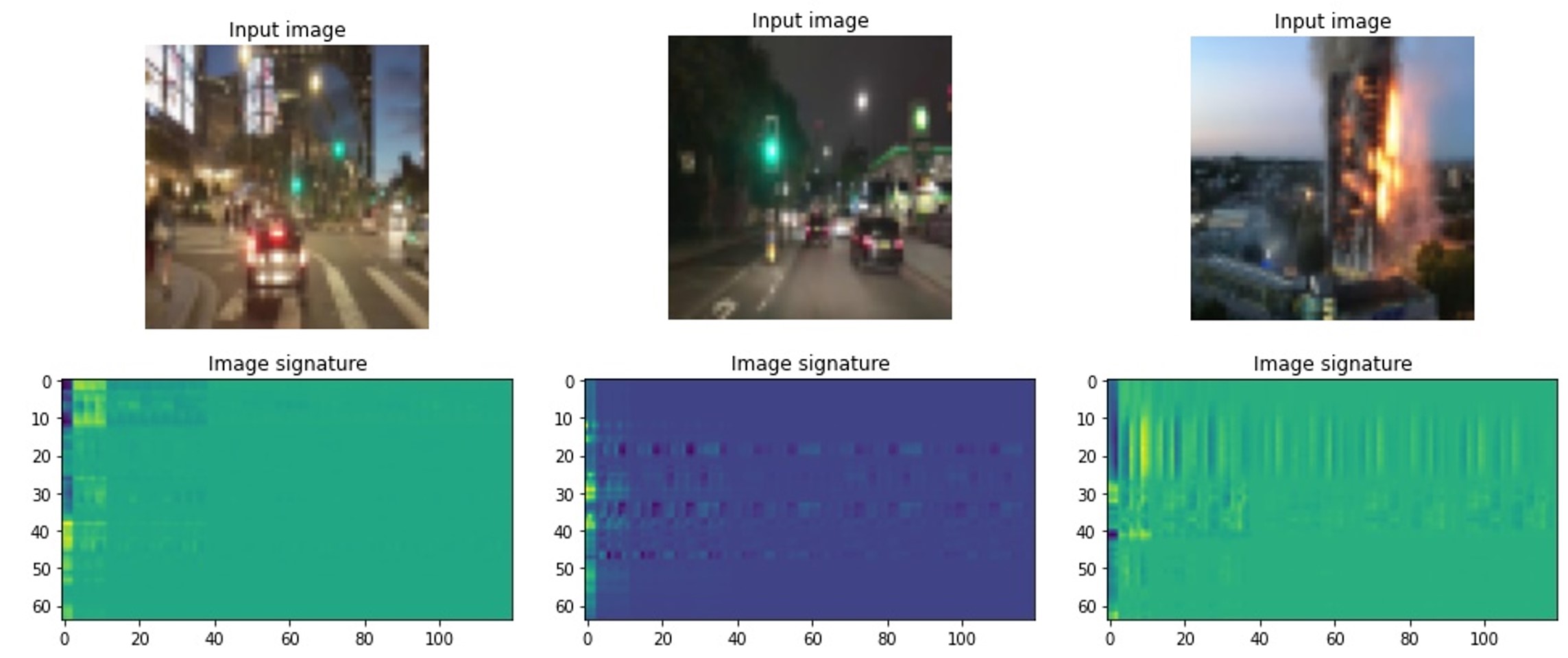}
\end{center}
   \caption{A sample of images from fire dataset (first row) and their low-dimensional representation as a unique signature (second row).}
\label{fig:sig_sample}
\end{figure}


\textbf{Fire dataset:} 
Finding a relatively large dataset that can represent anomaly detection such as fire detection from images remains a challenge. Accordingly, we have created our dataset that represents various conditions of the environment. The dataset comprises 18,912 images that are downloaded from the web, specifically Google Images. This dataset is used for training, validating and testing our method. We manually labelled the dataset to fire and no fire.

\textbf{Fog benchmark dataset \cite{weathernet}:} 
We also implemented several experiments for fog detection from street-level images, which is a crucial task for road safety. We used the dataset introduced in WeatherNet ensemble model. Also, we compare our results to their FogNet model. The dataset, for fog detection alone,  comprises 4,345 RGB images distributed into 718 and 3,627 for positive and negative classes, respectively.

\textbf{Concrete crack benchmark dataset \cite{crack_dataset,crack_detection_cnn}:} It comprises 40,000 RGB images distributed evenly between positive and negative cases for concrete crack detection. 

\textbf{CelebA benchmark dataset \cite{celeba}:} It comprises 202,599 RGB images of several facial attributes. We conducted several experiments for detecting some facial attributes (\eg wearing hat, bald, blond hair). 

In supplementary, we show the sample size per class and its weight in both training and testing sets for the four datasets. It is worth mentioning that the class weights represent ($\alpha$) values in the aforementioned focal loss function. 
During training and in absence of validation set, we have split the training set to train and validate sets in a ratio of 80-20\% to train and validate the models per each training cycle (epoch). During inference, we used the test set, which the models \emph{have never seen} during training or validation, for evaluating our pretrained models, computing metrics, and comparing their results. 

Fig. \ref{fig:sig_sample} shows a sample of images (from fire dataset) and their unique image signature (depth = 4), which shows how a given image can be translated to a stream of paths and projected to a lower dimension signature.

\subsection{Evaluations}
After training several models, we evaluated their pre-trained weights on the test-set for a given dataset, which the models have never seen during training and validation. Several metrics are computed to evaluate the trade-off between a given model’s performance and efficiency for computations and inference. Based on the aforementioned objective loss, we computed accuracy (acc), Average Precision (AP) and F1 score, which is defined as:  $F1 = 2 \cdot  Precision \cdot  Recall/(Precision + Recall)$. We also computed the Receiver operating characteristic (ROC) curve to evaluate the trade-off between true and false-positive rates.  

To reflect on a given model’s computational efficiency, we have calculated the number of parameters (params) for each model, its disk size, and the Floating-point Operations Per Second (FLOPS) for computing the operations of a given model in a batch size (1).  

We compared our results to state-of-the-art models (\eg ResNet50, VGG16, and MobileNet) to reflect on both: performance and efficiency for training and computation resources. We also trained a base model of a single FC layer on images directly to show the effectiveness of ImageSig with the same FC layer unequivocally.

\section{Results}

\begin{figure}
\begin{center}
\includegraphics[width=1\linewidth]{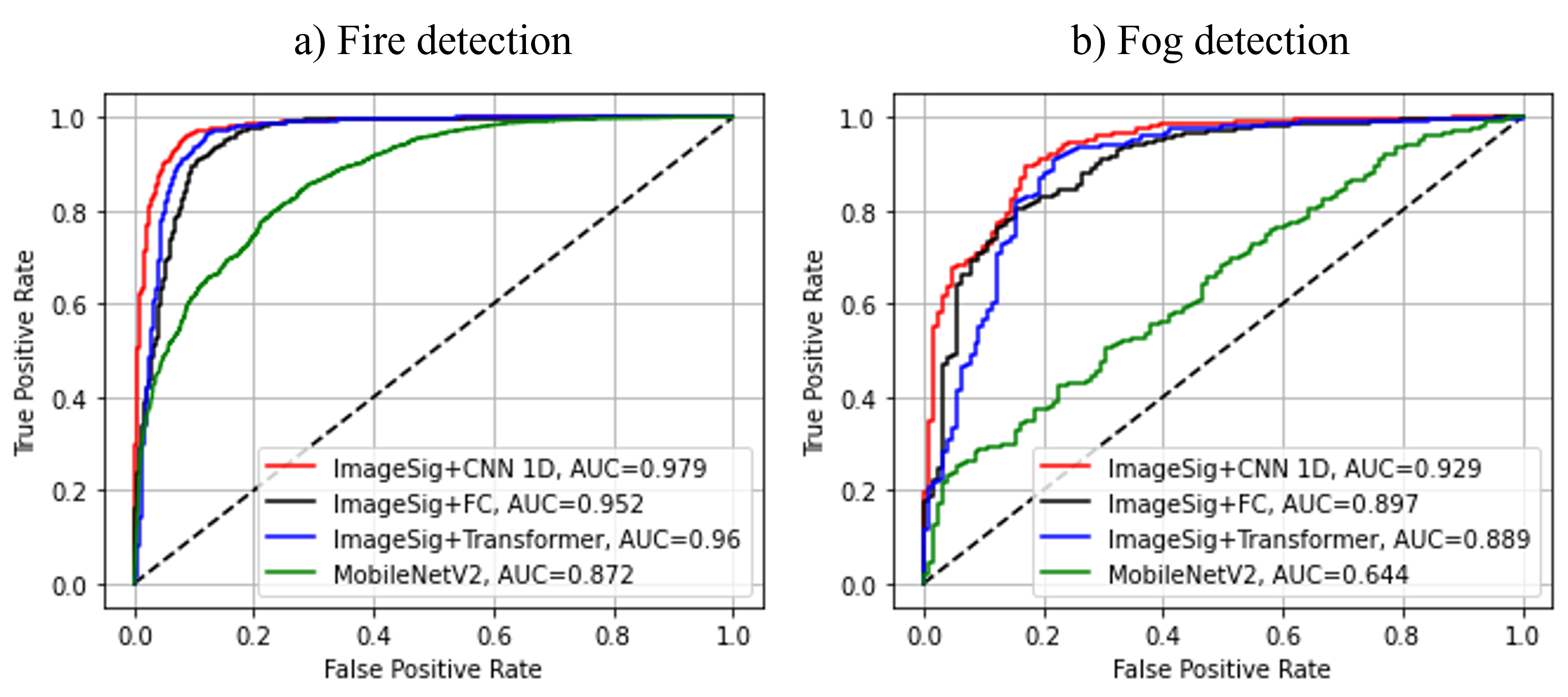}
\end{center}
   \caption{ROC curves for trained models on fire and fog datasets. The ImageSig models are trained on image resolution (64 X 64) with depth = 4, and batch (3000).}
\label{fig:roc}
\end{figure}


In Table \ref{tab:main_results}, we compare our methods with other state-of-the-art models when trained and tested on the fire dataset. It shows that our methods outperform other models when trained on the same image resolution (64,64). The table shows that ImageSig with a single FC layer is 3.6 times more accurate than relying only on a single FC layer without signature and achieves as high as VGG16 model, but with almost 50 times fewer parameters when trained on the same input size.  While a higher accuracy can be achieved when training, for instance, ResNet50 model on a larger image resolution, the introduced method outperforms all other methods when it comes to model’s efficiency, the number of parameters, training time, and FLOPS. In fact, a fully trained model of ImageSig + CNN1D has achieved the highest performance, for image resolution (64,64), with the lowest disk size (0.6 MB), fewer parameters (37,112) and FLOP of 1.69 million, showing the robustness and efficiency of the introduced method.

Table \ref{tab:main_results_fog_dataset} compares the results of ImageSig with a state-of-the-art model (FogNet) that relies on ResNet50. Similar to the previous results, while ResNet50 trained on a larger input can achieve higher accuracy, we show that ImageSig achieves a better performance in terms of AP and F1 scores with fewer parameters. 

For anomaly detection, understanding the relationship between true and false positive rates is crucial. Fig. \ref{fig:roc} shows the ROC curve for the introduced architectures of ImageSig, FC, CNN1D, Transformer, which shows a uniform relationship between true and false positive and high AUC values above 0.9 for fire and fog datasets.

Furthermore, we also compare our best results to other state-of-the-art models on the remaining two benchmarked datasets. Table \ref{tab:main_results_celebA_dataset} shows the results for detecting three facial attributes (wearing hat, bald and blond hair) on the test set of CelebA dataset. ImageSig shows competitive results in comparison to other deeper base models, bearing in mind the uneven trade-off between the difference in accuracy and the number of parameters for a given base model. Last, Table \ref{tab:main_results_crack_dataset} shows the results for detecting concrete crack, in which ImageSig has achievedd 98.7\% accuracy with an input size (64 X 64) and  at least 620 times fewer parameters when compared to the other base models.

\begin{table*}[ht]
\begin{center}
\begin{tabular}{l|c c c c c c c  c}
\hline
Method & acc (\%) & AP & F1 & Resolution & Params & Model size & Training time & FLOPS (M) \\
\hline\hline
Base model (FC)$^1$ & 25.54 & 0.74 & 0 & (64,64) & 614,552 & 7.05 MB & 4.32 min & 1.23 \\ 
\hline
ResNet50 \cite{ResNet50}& \textbf{99.23}  & - &  - & (224,224) & 23,591,810 & 270 MB & 82.2 min* & 7,751.21  \\
        & 91.97  & 0.90 & 0.94 & (64,64) & 23,591,810 & 90.5 MB & 6.88 min* & 632.90 \\
VGG16 \cite{vgg}   & 93.24  & 0.93 & 0.95& (64,64) & 14,714,688 & 56.2 MB & 7 min* & 2,507.18  \\
MobileNet \cite{mobilenets}& 81.68  & 0.80 & 0.88&(64,64) & 3,230,914 & 12.5 MB & 3.7 min* & 93.58\\
MobileNet V2 \cite{mobilenetv2}& 81.59  & 0.80 &0.88 &(64,64) & 2,260,546 & 9.10 MB & 4.63 min* & 50.11\\

\hline 
ViT \cite{vit} & 75.23 & 0.76& 0.85& (64,64) & 4,915,401 & 59.5 MB & 21.42 min &4.65    \\

\hline 
\textbf{ImageSig + FC} & 93.13   & 0.93 & 0.95 & (64,64) & 384,152 & 4.42 MB & 5.2 min & \textbf{0.77}\\  

\hline
\textbf{ImageSig + CNN1D} & 95.02  & 0.96 & 0.97 & \textbf{(64,64)} & \textbf{37,112} & \textbf{0.6 MB} & 5.4 min & 1.69\\  

\textbf{ImageSig + CNN1D} & 95.53  & 0.96 & 0.97 & (224,224) & 95,120 & 1.27 MB & 13.1 min & 6.17\\

\hline

\textbf{ImageSig + CNN1D$^2$} & \textbf{96.75}  & 0.97 & 0.98 & (64,64) & 59,512  & 0.9 MB & 14.3 min & 3.49\\
\hline
\textbf{ImageSig + Transformer} & 94.01  & 0.94 & 0.96 & (64,64) & 162,553 & 2.51 MB & 8.1 min & 1.79\\
\hline
\end{tabular}
\end{center}
\caption{Comparing the results of different methods on the test set of \textbf{fire dataset}. Both ImageSig models are trained from scratch with a signature depth = 4 and  with a batch size (3000). *All convolutional models (only) are trained on this dataset via transfer learning with ImageNet weights.
$^1$ Training the same FC layer architecture on images directly without signature.
$^2$It refers to computing signature for two directions of a given image.  Models are trained on a single GPU (Titan V).}

\label{tab:main_results} 
\end{table*}

\begin{table*}[ht]
\begin{center}
\begin{tabular}{l|c c c c c c c  c}
\hline
Method & acc (\%) & AP & F1 & Resolution  & Params & Model size & Training time & FLOPS (M) \\
\hline\hline
Base model (FC)$^1$ &16.51 &0.86 & 0 & (64,64) & 614,552 & 7.05 MB & 1.1 min & 1.23 \\  
\hline
FogNet-ResNet50 \cite{weathernet} & \textbf{95.60}  & 0.86 &  0.84  & (224,224) & 23,591,810 & 270 MB & - & 7,751.21 \\

\hline 

\textbf{ImageSig + FC} & 87.38 & 0.93 & 0.92 & (64,64) & 384,152 & 4.42 MB & \textbf{0.84 min} & \textbf{0.77}\\

\textbf{ImageSig + CNN1D} & 88.50  & \textbf{0.95} & \textbf{0.93} & \textbf{(64,64}) & \textbf{37,112} & \textbf{0.6 MB} & 1.2 min & 1.69\\
\textbf{ImageSig + Transformer} & 90.17  &  0.86 &0.89 & (64,64) & 162,553 & 2.51 MB &  2.1 min & 1.79\\
\hline
\end{tabular}
\end{center}
\caption{Comparing the results of different methods on the test set of \textbf{fog dataset}. All ImageSig models are trained similar to the ones trained for fire detection without any additional fine-tuning. $^1$ Training the same FC layer architecture on images directly without signature.
}
\label{tab:main_results_fog_dataset} 
\end{table*}

\begin{table}[ht]
\begin{center}
\begin{tabular}{l|c c c c   }
\hline
Method & W. Hat & Bald & Blond & Params  \\

     & acc (\%)  & acc (\%)  &acc (\%)  & (M)  \\
\hline\hline

FaceTracer \cite{FaceTracer} $^*$  & 89 & 89 & 80 & - \\
PANDA-w \cite{PANDA} & 91 & 92 & 81 & $>$ 2 \\
PANDA-l\cite{PANDA} $^*$ & 96 & 96 & 93  & $>$ 2\\
\cite{SurfCascade}+ANet \cite{celeba} & 93 & 92 & 86  & $>$ 61\\
LNets+ANet$^1$  & 96 &95 & 91 & $>$ 61 \\
LNets+ANet   &  \textbf{99} & \textbf{98}  &\textbf{95}  & $>$ 61  \\
\hline

\textbf{ImageSig$^2$}& 91 & 92 & 84 & \textbf{0.037} \\
\hline
\end{tabular}
\end{center}
\caption{ Comparing the results of different models on the test set of \textbf{CelebA dataset}. The results of base models are obtained from \cite{celeba}. ANet is based on AlexNet, with estimated params $\approx$ 61 M. $^1$refers to "without pretraining". $^*$Models obtain facial parts based on landmark points. $^2$ Refers to ImageSig + CNN1D, trained similar to the models for the aforementioned datasets without any special fine-tuning.}
\label{tab:main_results_celebA_dataset} 
\end{table}

\begin{table}[ht]
\begin{center}
\begin{tabular}{l|c c c  c c c c}
\hline
Method & acc (\%)  & F1 & Resolution  & Params (M) \\
\hline\hline

ResNet152 & 99.5  & \textbf{0.99} & (224,224) & 60.2  \\
ResNet101 & \textbf{99.9}   & \textbf{0.99} & (224,224) & 44.5 \\
ResNet50 & \textbf{99.9} & \textbf{0.99} & (224,224) & 25.6 \\

\hline

\textbf{ImageSig$^1$}& 98.7 & \textbf{0.99} & \textbf{(64,64}) & \textbf{0.037}\\
\hline
\end{tabular}
\end{center}
\caption{ Comparing the results of different models on the test set of \textbf{Concrete crack dataset}. The results of ResNet models are obtained from \cite{crack_detection_cnn}. $^1$ Refers to ImageSig + CNN1D, trained similar to the models for the aforementioned datasets.}
\label{tab:main_results_crack_dataset} 
\end{table}

\begin{figure}
\begin{center}
\includegraphics[width=0.7\linewidth]{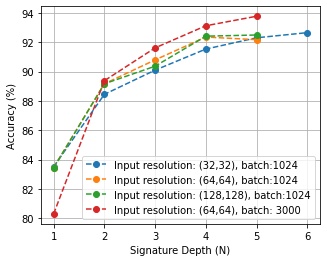}
\end{center}
   \caption{The relationship between image resolution, signature depth, batch size and model performance for ImageSig + FC models, neurons= 50. }
\label{fig:depth_acc}
\end{figure}

\section{Ablation studies}
We trained several models with different hyperparameters to evaluate the performance of the introduced method. All models are trained on the fire dataset with only a single FC layer after flattening the computed signatures. The aims of these studies are to understand the effect of the input size, signature depth and the type of signature on both accuracy and number of parameters and subsequently the memory footprint for a given model. Fig. \ref{fig:depth_acc} shows the relationship between image resolution, signature depth and the overall performance of a given model.

\begin{table*}
\begin{center}
\begin{tabular}{l| c c c c c c}
\hline
Resolution & Depth (N) & acc (\%)  & Params & Model size & Training time & FLOPS (M)\\
\hline\hline

(32,32) & 1 & 83.51 & \textbf{4,952} & \textbf{84 KB} & \textbf{1.43 min} & \textbf{0.01} \\
        & 2 & 88.45	& 19,352 &253 KB& 1.21 min & 0.04 \\
        & 3 & 90.10 & 62,552 &759 KB& 1.82 min & 0.12 \\
        & 4	& 91.54 & 192,152 &2,278 KB & 4.16 min & 0.38 \\
        & 5	& \textbf{92.31} & 580,952 &6,834 KB & 10.17 min & 1.16 \\
        & 6	& \textbf{92.66}	& 1,747,352	& 20,994 KB	&16.34 min & 3.49 \\
\hline
(64,64) & 1 & 83.45 &	9,752 &140 KB &1.54 min & 0.02 \\
        & 2 & 89.17 &	38,552 &478 KB&	1.94 min & 0.08\\
        & 3 & 90.80 &	124,952	&1,490 KB&	3.21 min & 0.25 \\
        & 4 & \textbf{92.36} &	384,152& \textbf{4,528 KB} &	5.22 min & \textbf{0.77} \\
        & 5 & 92.18 &	1,161,752&	13,640 KB&	17.42 min & 2.32 \\
\hline
(128,128) & 1 & 83.43&  19,352&	258.5 KB&	1.52 min  & 0.04 \\
        & 2 & 89.17&	76,952&	949.7 KB&	1.30 min  & 0.15\\
        & 3 & 90.37&	249,752&	3,023 KB&	2.81 min & 0.50 \\
        & 4 & \textbf{92.42}&	768,152&	9,244 KB&	7.46 min  & 1.53 \\
        & 5 & \textbf{92.50}&	2,323,352&	27,906KB&	22.26min & 4.65 \\
\hline
\end{tabular}
\end{center}
\caption{Comparing the effects of different image resolutions and signature depth on the performance. All models are trained with a single FC layer of 50 neurons for 200 epochs on fire dataset with a batch size \textbf{(1024)}.}
\label{tab:comparing_input}
\end{table*}

\textbf{Images as two directional streams:} 
Table \ref{tab:main_results} also shows the effect of modelling images as streams by computing and combining signatures for both directions (horizontal and vertical paths) on the overall generalization of ImageSig. It shows an increase of 1.73 \% in accuracy when training ImageSig + CNN1D on fire dataset. However, this increase comes at the expense of doubling FLOPS (from 1.69 million to 3.49 million).

\textbf{The effect of signature depth:}
The depth of the truncated signature directly affects the performance of a given model. Table \ref{tab:comparing_input} shows the results of 16 models of different image resolutions and signature depths. For a given image resolution, the increase in signature depth leads to a direct increase in the performance of a given model in a non-linear relationship, whereas an increase in depth after certain points could exacerbate the performance of the model. On the other hand, it is worth mentioning that this increase in depths (within the linear part of the relationship) comes at the expense of the number of parameters and consequently the training time and the model size. Generally, for this given classification task, it seems that a depth of 4 provides the top score in relation to the number of parameters and the model’s disk size. 

\textbf{The effect of the input size:}
While the effect of the input size on the performance of a given model is less significant when it is compared to the signature depth, a larger input size seems to provide a trained model with better accuracy at a shallower depth of a signature. This also benefits the trade-off between performance and the number of parameters. For example, in Table \ref{tab:comparing_input}, a model of resolution (32 X 32) at depth 5 achieves a similar accuracy for a model of resolution (64 X 64) at depth 4, whereas the number of parameters in the latter model is 1.5 less than the aforementioned one. Accordingly, a smaller input size does not necessarily mean less number of parameters, but the trade-off between the input size and the depth of computed signature requires optimisation for a given task.

\begin{table}[ht]
\begin{center}
\begin{tabular}{l| c c c c  }
\hline
Resolution & Sig. type  & acc (\%)  & Params & Model size  \\
\hline\hline

(32,32), & Sig. &	92.31&	580,952&	6,834 KB  \\
 N = 5                  & Log Sig. & 89.04&	128,152& 1,582 KB \\

\hline
(64,64),  & Sig. &	92.36&	384,152&	4,528 KB\\
 N = 4                 & Log Sig. & 88.91&	102,552&	1,228 KB \\
\hline
\end{tabular}
\end{center}
\caption{Comparing the effects of the type of image signature on the performance. All models are trained with with a single FC layer with 50 neurons.}
\label{tab:sig_type}
\end{table}

\textbf{Signature vs logarithmic of signature:} 
While the logarithmic version of signature leads to normalisation and dimensionality reduction of a given input, we found that the linear version of signature achieves better performance for a given model. Table \ref{tab:sig_type} shows the results of four models at two different image resolutions and signature depth, in which we have changed only the type of signature for each image resolution. The table shows that the signature outperforms the log signature for both image resolutions at different depths. However, the log signature can provide models with fewer parameters and less model size and training time.

\textbf{Post-training quantization :}
Table \ref{tab:quant} shows the results of model size before and after post-training dynamic range quantization  for state-of-the-art and imageSig methods. It shows unprecedentedly that a full vision model can reach a few kilobytes (44.2 KB) which is suitable for edge devices with limited memory and computational resources. It also shows that even after compressing the state-of-the-art methods, their size remains larger than ImageSig models without quantization. 

\begin{table}
\begin{center}
\begin{tabular}{l| c c}
\hline
Model & Model size & Quantized size \\
\hline\hline

ResNet 50 & 90.5 MB & 23 MB  \\
MobileNet & 12.5 MB  & 3.30 MB\\
MobileNetv2 & 9.10 MB &2.51 MB \\
\hline
ViT & 59.5 MB & 8.88 MB \\
\hline
ImageSig + FC & 4.42 MB &  377 KB \\
\textbf{ImageSig + CNN1D} & \textbf{0.6 MB} & \textbf{44.2 KB}  \\
ImageSig + CNN1D$^1$ &0.9 MB& 66.1 KB  \\

\hline
\end{tabular}
\end{center}
\caption{Model size: Before and after quantization}
\label{tab:quant}
\end{table}

\begin{table}[ht]
\begin{center}
\begin{tabular}{l| c c  c}
\hline
Method &Hardware &FPS& FPS$^*$  \\

\hline\hline

MobileNetV2  &RaspberryPi4$^2$ &  13.9  &7.4 \\
                &Jetson-nano$^3$ & 46.9 &19.8  \\
                & RTX 2080ti & 62.2 & 37.27\\
\hline
ImageSig + CNN1D &RaspberryPi4$^2$ & \textbf{142.3} &  \textbf{15.4}   \\
                &Jetson-nano$^3$ & \textbf{442.8} &  \textbf{24.3} \\
                & RTX 2080ti & \textbf{1008.4} & \textbf{90.69} \\
\hline
\end{tabular}
\end{center}
\caption{Performance of AI at the Edge. All tested models are trained on fire dataset and converted to tflite. $^1$ model with resolution (64,64) and depth = 4, trained on fire dataset. $^2$Raspberry pi CPU arm71 linux with 4 GB Ram. $^3$Jetson nano includes a GPU and 4 GB Ram. $^*$ This metric includes data structuring and prepossessing, in addition to the model inference. }
\label{tab:AI_performance}
\end{table}
\textbf{Embedded AI at the edge:}
Table \ref{tab:AI_performance} shows that our methods outperform MobileNetV2 when it comes to efficiency, in terms of Frame Per Second (FPS), for embedded AI at the edge. It shows that our method achieves an unprecedented 15.4 FPS on Raspberry Pi v4 (CPU) for a vision model including all necessary data structuring and preprocessing and 142.3 FPS when considering only the performance of the model. It also shows a real-time performance that outperforms MobileNetV2 for Jetson nano with GPU acceleration.

\emph{Implementation details and further studies regarding performance, number of neurons in the FC layer and batch size are addressed in the supplementary section.}


\section{Remarks and future work}
We introduced a new method, called ImageSig, for ultra-lightweight image recognition that has been tested on several datasets without any \emph{special fine-tuning} for a given task.  ImageSig relies on extracting a unique signature of a given image that can be feedforward to a single FC layer, CNN 1D, or a transformer for classification. The key advantage of ImageSig is training efficient models that can be embedded in real-world devices with limited computational resources and memory footprint (\eg smart cameras, car sensors, or fire alarms). With highly accurate models and less than a million FLOPS, ImageSig method could allow diversifying the applications of AI at the edge. The second key advantage is its low impact on the environment; a full model can be trained in a few minutes or less, minimising the power consumption and CO2 emission when compared to other state-of-the-art methods of similar performance. As for future research, we aim to extend ImageSig to other vision tasks (\eg object detection and semantic segmentation), in which further research is required to consider both local and global signatures.

\section{Acknowledgments}

This work was supported in part by the EPSRC [grant number EP/S026347/1], in part by The Alan Turing Institute under the EPSRC grant EP/N510129/1, in part by The Alan Turing Institute’s Data Centric Engineering Programme under the Lloyd’s Register Foundation grant G0095, in part by The Alan Turing Institute’s Defence and Security Programme, funded by the UK Government, in part by The Alan Turing Institute’s Office of National Statistics Programme, funded by the UK Government and in part by the Hong Kong Innovation and Technology Commission (InnoHK Project CIMDA).

{\small
\bibliographystyle{unsrt}

\bibliography{egbib}

\section{Supplementary}

\subsection{ImageSig as a python module}

Here is an example for training and inference using ImageSig as a python module. 

\begin{verbatim}


### Training 

from imagesig import image_signatrue
SIG_DEPTH = 4
IMAGE_SIZE = (64,64) 
TRAIN_DIR = "data/training_set"
TEST_DIR = "data/test_set

train_sig = image_signature (
            image_dir = TRAIN_DIR,
            depth = SIG_DEPTH,
            image_size = IMAGE_SIZE,
            flatten=FLATTEN,
            log_sig = False,
            two_direction = True,  
            augment_flip_horizontal = True,
            augment_flip_vertical =True,
            augment_brightness= True
            )
test_sig = image_signature (
            image_dir = TEST_DIR,
            depth = SIG_DEPTH,
            image_size = IMAGE_SIZE,
            flatten=FLATTEN,
            log_sig = False,
            two_direction=True
            )
_,x_train,y_train = train_sig.read_dir()
_,x_test,y_test = test_sig.read_dir()

......  Model architecture ......


### Inference

from imagesig import imagesig_predict
model = "checkpoint"
image = "fire.jpg"
pred = imagesig_predict(image,model)

\end{verbatim}

\subsection{Dataset details and accessibility}

Table \ref{tab:data_fire} shows how the image samples are distributed per each class in training and testing for all datasets. Test set is the data subset where all models are evaluated, in which the model never seen during training and validation of each epoch.

\textbf{Fire dataset access:}
To be made available upon request.

\textbf{Fog dataset access:}
Can be requested from WeatherNet's authors:
\url{https://www.mdpi.com/2220-9964/8/12/549}

\textbf{Concrete crack dataset access:}
\url{https://data.mendeley.com/datasets/5y9wdsg2zt/2}

\textbf{CelebA dataset access:}
\url{https://mmlab.ie.cuhk.edu.hk/projects/CelebA.html}

\begin{table}[ht]
\begin{center}
\begin{tabular}{l| c c c c}
\hline
Dataset & Subset & Classes  & Sample  &  $(\alpha)$ \\
\hline\hline
Fire & Training & Fire & 4,097 & 1.84 \\
        & & No fire & 11,055  & 0.68 \\
 & Test  & Fire & 960  & - \\
      &   & No fire & 2,800 & - \\
\hline

Fog & Training & Fog & 589 & 2.93 \\
        & & No fog & 2,860  & 0.60 \\
 & Test  & Fog & 129  & - \\
      &   & No fog & 769 & - \\
      
\hline

Concrete  & Training & Negative & 16,000 & 1.00 \\
crack       & & Positive &  16,000 & 1.00 \\
 & Test  & Negative & 4,000  & - \\
      &   & Positive & 4,000 & - \\

\hline

CelebA  & Training  & Multi-task & 162,770 & -\\
 & Validation  & Multi-task & 19,867 & - \\
 & Test  &Multi-task & 19,962 & - \\

\hline
\end{tabular}
\end{center}
\caption{Data sample distribution and class weight.}
\label{tab:data_fire}
\end{table}
\subsection{Implementation details}
\textbf{All codes are provided} for our implementations in the ZIP file. Here is a brief description of our implementations:

\textbf{Image signature:} 
We have created our own python class for implementing and augmenting signature to images as paths of streams, as described in this paper, called image signature. We built image signature based on computing signature from two different python packages as backends for image signature so-called Signatory \cite{kidger2021signatory} and iisignature \cite{iisignature}. We did not observe a noticeable effect on which backend we used on the performance of a given model. Accordingly, We relied mainly on Signatory for all models, however, for edge devices such as RaspberryPi we used iisignature, given that signatory depends on Pytorch for acceleration whereas iisignature outputs numpy arrays that can fit both tensorflow and Pytorch models without the need for converting torch tensors to TF tensors. The main models are computed with input resolution of 64 X 64 and signature depth = 4. However, in the ablation studies, we show the effect of various signature depth and image input size on the performance of the model. 

\textbf{Data augmentations:} 
All models are trained with the same data augmentation techniques such as normalizations,  horizontal flipping, and changing colour brightness.

Fig. \ref{fig:data_augmentation} shows an example for the augmented images and their respective signatures for training.

\begin{figure*}
\begin{center}
\includegraphics[width=1\linewidth]{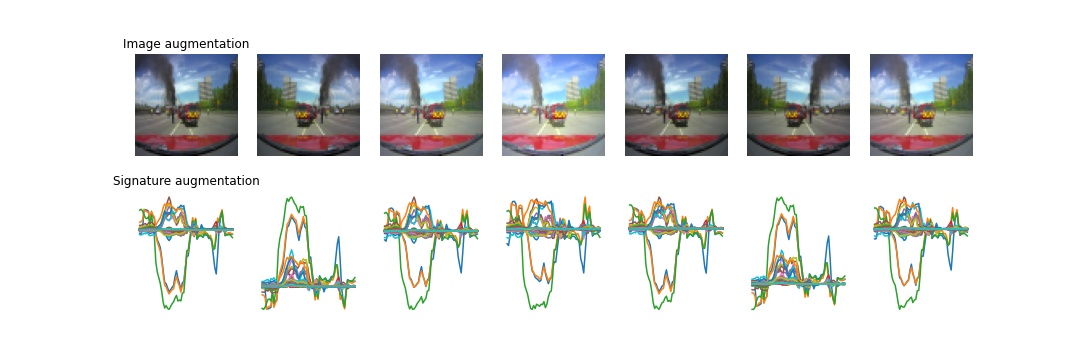}
\end{center}
   \caption{Sample of signature augmentations.}
\label{fig:data_augmentation}
\end{figure*}


\textbf{ImageSig FC-based Mode:} 
We trained several models with different hyper-parameters that only rely on a  single FC layer. This FC layer comprises 50 neurons and is activated by a ReLU function. Models are trained with Adam optimiser of batch size 3000 and for 300 epochs. In ablation studies, we show the effect of the selected different hyperparameters on the performance of the model such as changing the number of neurons or batch size. 

\textbf{ImageSig convolution-based model:} 
After computing signature, we used a simple convolution block comprised of two CNN 1D layers of feature maps of 32 and 64 respectively and a kernel size of 3, in which both layers are activated by a ReLU function. Each layer is followed by a Max-pooling layer of kernel size 3. After the last pooling layer, the model is flattened and feedforward to a single FC layer of 50 neurons that is activated by a ReLU function before the final softmax layer. The model is trained based on Adam optimisation \cite{adam} and a batch size of 3000 for 300 epochs. 

\textbf{ImageSig Attention-based model:} 
after the signature input, we added a convolution block implemented similar to the aforementioned architecture. After the convolution block, we normalized the final output of the last pooling layer and passed it to a multi-head attention layer of eight heads. Afterwards, we add a skip connection that connect the attention outputs and the pooling layer. For each transformer layer (50 in total), we added a normalization and a MLP layers similar to transformer units [64,64]. Finally, we added an FC layer of 50 neurons. The model is trained with 64 projection dimensions that is equivalent to dimension of a given input. All layers are activated by a ReLU function, except the softmax output layer. The model is trained with AdamW optimiser with a learning rate of 0.001, and a decay of 0.0001 with a batch-size of 300 and 3000 epochs.

\textbf{Convolution base models:} 
All image convolution models are implemented via transfer learning on the given custom dataset with ImageNet weights. After truncated the models, we trained an FC and output layers with the same hyperparameters of the aforementioned architectures. All models have converged when trained for 20 epochs and patch size 1024.  

\textbf{ViT base models:} 
We trained transformer models from scratch on the mentioned datasets. We followed the implementation details for ViT model in their original paper closely. The model is optimised based on AdamW and patch size of 1024 and trained for 300 epochs.

\subsection{Further Results}

From Fig. 2-7, we show a sample of predictions of our models on the different datasets showing true and false prediction.

\begin{figure}
\begin{center}
\includegraphics[width=0.8\linewidth]{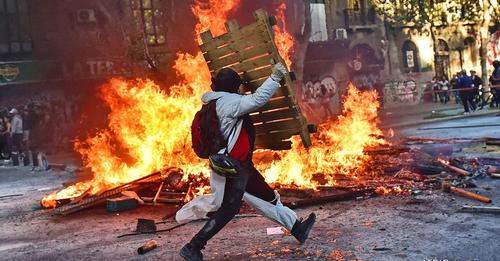}
\end{center}
   \caption{Predicted values: fire, no fog.}
\label{fig:result_sample}
\end{figure}


\begin{figure}
\begin{center}
\includegraphics[width=0.8\linewidth]{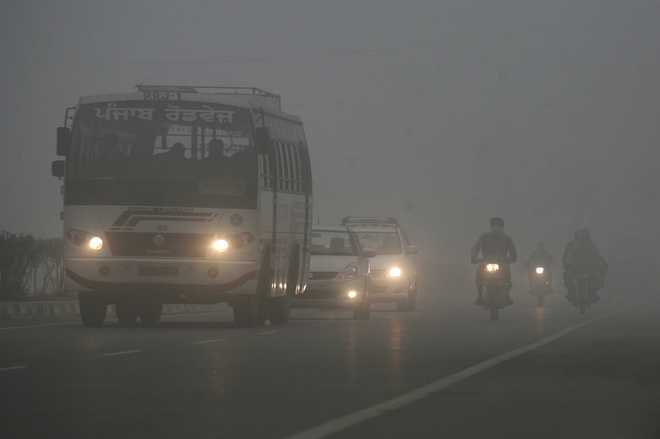}
\end{center}
   \caption{Predicted values: No fire, fog.}
\label{fig:result_sample2}
\end{figure}


\begin{figure}
\begin{center}
\includegraphics[width=1\linewidth]{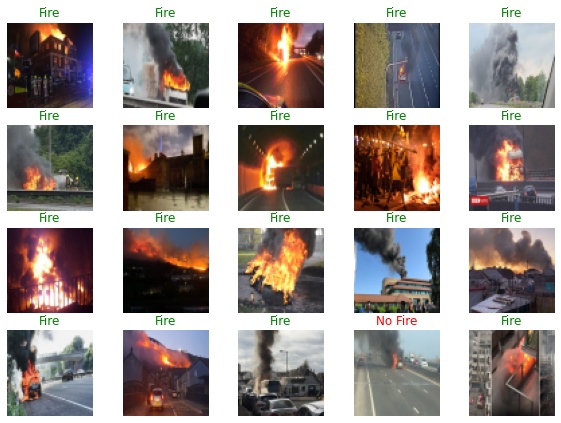}
\end{center}
   \caption{A sample of prediction on the fire dataset. Model predictions (green: correct, red: incorrect). }
\label{fig:sample_fire}
\end{figure}


\begin{figure}
\begin{center}
\includegraphics[width=0.9\linewidth]{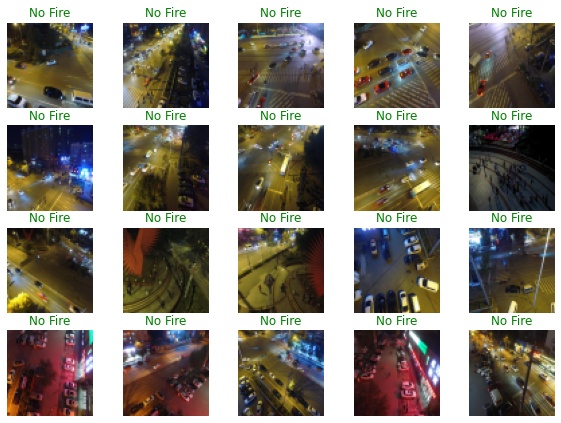}
\end{center}
   \caption{A sample of prediction on the fire dataset. Model predictions (green: correct, red: incorrect). }
\label{fig:sample_nofire}
\end{figure}


\begin{figure}
\begin{center}
\includegraphics[width=0.9\linewidth]{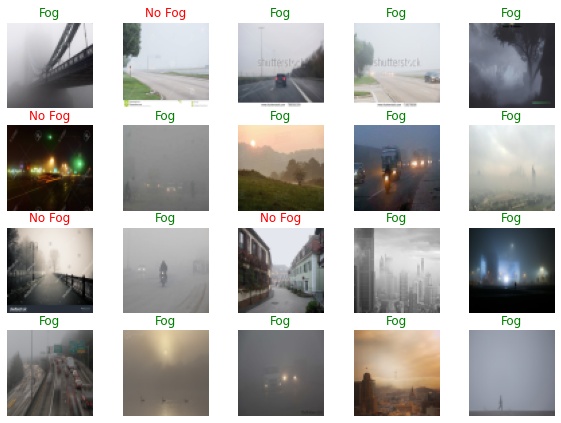}
\end{center}
   \caption{A sample of prediction on the fog dataset. Model predictions (green: correct, red: incorrect). }
\label{fig:sample_fog}
\end{figure}


\begin{figure}
\begin{center}
\includegraphics[width=0.9\linewidth]{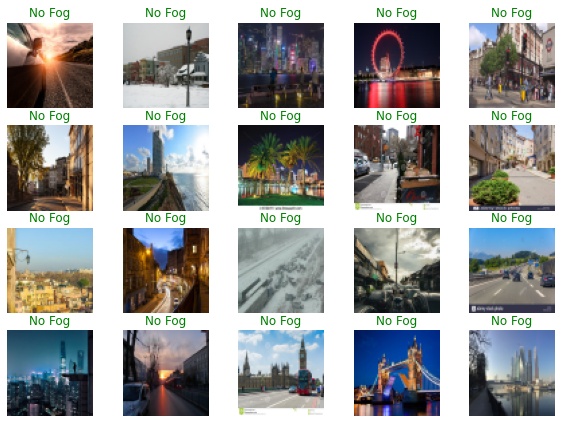}
\end{center}
   \caption{A sample of prediction on the fog dataset. Model predictions (green: correct, red: incorrect). }
\label{fig:sample_nofog}
\end{figure}


\subsection{Further ablation studies}

\textbf{The effect of the number of neurons:}

We noticed a direct effect of the number of neurons of the fully connected layer on the performance of a given model and its number of parameters. Table \ref{tab:num_neurons} summarises the results of 10 models to investigate the effect of the number of neurons. All models are trained with a single fully connected layer of different numbers of neurons. For every 5 models, the image resolution and the depth of the signature are kept constant, whereas the number of neurons has been changed between 25-200 neurons. The table shows that, at a given image resolution and signature depth,  by doubling the number of neurons,  a slight increase $(<1\%)$ in the accuracy can be achieved, bearing in mind the double increase of the mode’s parameters and consequently its disk size. We found that optimal results as a trade-off between the accuracy and model size can be achieved within the choice of 50 neurons as a hyperparameter of the fully connected layer.

\begin{table*}[ht]
\begin{center}
\begin{tabular}{l| c c c c c }
\hline
Resolution & Neurons  & acc (\%)  & Params & Model size & Training time  \\
\hline\hline

(32,32), Depth = 4 & 25 &	90.53&	96,072&	1,180 KB&	2.34 min  \\
                   & 50 & 91.54&	192,152& 2,278 KB& 4.16 min  \\
                   & 100 &92.23&	384,302& 4,638 KB&	2.38 min *  \\
                   & 150 &92.10&	576,452& 6,945 KB&	2.35 min *  \\
                   & 200 &93.00&	768,602& 9,253 KB&	2.35 min *  \\

\hline
(64,64), Depth = 4 & 25 &	91.19&	192,077& 2,332 KB&	3.99 min*  \\
                   & 50 &   92.36&	384,152& 4,528 KB& 5.22 min \\
                   & 100 &92.39&	768,302& 9,246 KB&	4.02 min*  \\
                   & 150 &91.99&	1,151,452&	13,857 KB& 4.02 min* \\
                   & 200 &92.87&	1,536,602&	18,469 KB&	4.04 min* \\
\hline
\end{tabular}
\end{center}
\caption{Comparing the effects of the number of neurons of the single FC layer on the performance.All models are trained with a fixed signature depth (N=4).}
\label{tab:num_neurons}
\end{table*}

\textbf{The effect of batch size: }

In table \ref{tab:batchsize}, we show the effect of  batch size on accuracy. From our experiments, larger batch size allows stability in training and convergence. Giving that the model requires very small memory, training it with a large batch size can be achieved with a minimal shared memory in comparison to other models such as ResNet or MobieNetV2.

Furthermore, Table \ref{tab:comparing_batchsize} shows the effect of the batch size on the overall performance of a given model in relation to the signature depth. The table shows a better results at a batch size of 3000 when compared to a batch size of 1024, for the same image resolution and signature depth.

\begin{table}[ht]
\begin{center}
\begin{tabular}{l| c c c c c}
\hline
Batch size &64 &128 &256 &512 &1024   \\
\hline\hline

acc (\%)  &79.01 & 88.13  & 90.36 & 92.52 & 94.38 \\

\hline
\end{tabular}
\end{center}
\caption{The effect of batch size on acc on fire dataset}
\label{tab:batchsize} 
\end{table}

\begin{table}[ht]
\begin{center}
\begin{tabular}{l| c c c}
\hline
Batch size & Sig. Depth & acc (\%) & Training time (\%) \\
\hline\hline

1024 & 1 & 83.45  &1.54 min\\
        & 2 & 89.17 &1.94 min \\
        & 3 & 90.80 &	3.21 min\\
        & 4 & \textbf{92.36} &	5.22 min \\
        & 5 & 92.18 &	17.42 min \\
\hline
3000 & 1 & 80.29& 	0.72 min \\
        & 2 & 89.38 &	1.02 min \\
        & 3 & 91.62&	1.88 min \\
        & 4 & 93.40&	4.47 min  \\
        & 5 & \textbf{93.78}&	12.23 min  \\
\hline
\end{tabular}
\end{center}
\caption{Comparing the effects of different batch size on the performance. All models are trained with a single FC layer of 50 neurons for 200 epochs on fire recognition dataset.}
\label{tab:comparing_batchsize}
\end{table}

\end{document}